# TOWARD MACHINE RISK PERCEPTION: INTEGRATING TRUST CALIBRATION AND PRECURSOR-BASED RISK ESTIMATION FOR HUMANOID

**He Wen**
AI Safety Lab, Bailey College of Engineering & Technology, Indiana State University,
Terre Haute, IN

## ABSTRACT

*Humanoid robots are emerging as co-workers in smart manufacturing, yet their dynamic, human-like movements introduce safety risks that differ fundamentally from those of fixed or wheeled robots. Conventional safety paradigms based on reactive force or distance limits fail to capture the sequential, uncertain nature of humanoid failures. This study proposes a precursor-driven, trust-calibrated framework to enable proactive humanoid risk perception. Accident evolution is modeled through sequential precursor cues using a Logistic–Exponential (LE) formulation that couples logistic escalation from diverse precursors with exponential decay for temporal dissipation. Trust is defined as the inverse of the estimated accident probability, allowing humanoids to adapt behavior in real time, reducing aggressiveness when risk intensifies, and restoring confidence as stability returns. A multi-source dataset of 126 documented events and 241 precursors revealed twelve dominant accident modes, most evolving through overlapping cues within one second. A simulated case study ("fall-onto-human") demonstrated how the LE–Trust coupling can trigger early intervention and prevent collapse. The results advance humanoid safety from static thresholds toward dynamic, evidence-based inference, establishing a foundation for risk-aware and trustworthy human–robot collaboration in Industry 5.0 environments.*

## 1. INTRODUCTION

The rapid emergence of humanoid robots in industrial environments signals a paradigm shift in smart manufacturing [1]. Unlike conventional collaborative robots that operate within constrained kinematic envelopes, humanoids combine bipedal locomotion, whole-body dynamic control, and dexterous manipulation [2,3]. This morphology enables them to navigate human-designed workspaces, perform semi-structured tasks, and collaborate with workers in roles traditionally reserved for humans, such as part retrieval, bin holding, tool handover, and ergonomic load positioning [4]. As manufacturers seek greater flexibility, rapid reconfiguration, and labor augmentation, humanoids are becoming viable co-workers on assembly lines, logistics zones, and maintenance environments.

However, humanoid deployment also introduces a fundamentally different risk landscape [5,6]. Traditional industrial robot safety frameworks assume rigid mechanical bases, deterministic motion profiles, and safeguarded work envelopes [7,8]. Collaborative robots (cobots) extend this paradigm by limiting speed and force, supporting compliant interaction, and incorporating human presence sensing [9,10]. Yet these protections are insufficient for humanoids. Their ability to walk, bend, regain balance, and perform fast momentum-driven recovery movements generates failure modes, such as slips, torso-rotation impacts, leg-swing collisions, and object drops, that have no equivalents in fixed-base systems [6,11]. Safety challenges are amplified by real-world uncertainty: imperfect perception, occlusions, human unpredictability, and system latency can produce rapid shifts from nominal behavior to unstable, high-energy motion [12].

Current robotic safety methods predominantly rely on reactive constraints, including force limits, emergency stops, or predefined safety zones [7,9]. These mechanisms assume that hazards manifest abruptly and can be addressed by halting motion upon threshold violation. In contrast, human safety behavior is anticipatory; workers monitor subtle precursor cues such as balance shifts, hesitation, shaky grip, gaze diversion, posture changes, or environmental clutter to foresee danger and adjust behavior proactively [13–15]. To operate as safe industrial teammates, humanoids must similarly perceive weak signals, update internal confidence, and intervene before a hazardous contact, fall, or misjudgment materializes [16,17]. At the very least, they must not harm humans. This requires them to have a certain level of risk perception capability, which can be rule-based or data-driven, and even human empathy to strengthen mutual trust.

Therefore, two pressing frontier issues need to be addressed: machine risk perception and trust calibration [18–20]. In this work, machine risk perception refers to a system's capability to infer the likelihood of hazardous outcomes from evolving precursor cues under uncertainty. Trust calibration denotes the dynamic adjustment of a machine's internal confidence and behavior in response to perceived risk, avoiding both overconfidence and excessive conservatism.

Specifically, machine risk perception has traditionally been grounded in industrial robot safety standards (e.g., ISO 10218, ISO/TS 15066), which emphasize reactive safeguards such as force/velocity limits and speed-and-separation monitoring [21]. More recent human-robot collaboration (HRC) research moves toward proactive approaches, leveraging perception, human-motion forecasting, and probabilistic planning to anticipate contact rather than merely respond to threshold violations [22–24]. Safety-critical control, chance-constrained planning, and predictive safety filters further formalize risk-aware autonomy under uncertainty [25,26], while the National Institute of Standards and Technology (NIST) guidance extends dynamic protective-distance computation for factory deployment. Human-factors studies underscore trust calibration as central to safe collaboration, highlighting the need to adjust behavior under uncertainty rather than maintain static assurance [27–29]. However, these advances predominantly concern fixed-base or wheeled platforms; humanoids introduce unique hazards, such as fall dynamics, balance recovery, grip instability, and occlusion-driven perception lapses, that require interpreting weak precursor cues and modulating internal confidence continuously. This gap motivates a precursor-driven, trust-calibrated safety framework tailored to humanoid behavior in shared workspaces.

Therefore, this paper introduces a precursor-driven, trust-calibrated framework that enables humanoids to perceive and anticipate emerging risks proactively. The framework models hazard evolution through sequential precursor cues, estimates accident probability via a Logistic–Exponential (LE) process, and dynamically adjusts internal trust as the inverse of risk. This trust-guided mechanism allows humanoids to modulate motion and interaction behaviors adaptively, promoting cautious operation under rising uncertainty and restoring confidence as stability returns. The main contributions are as follows:

i. Development of a unified accident–precursor taxonomy for humanoid collaboration in industrial settings.

ii. Formulation of a dynamic trust model reflecting machine confidence under perceptual, behavioral, and control uncertainty.

iii. Integration of a logistic–exponential probabilistic model to capture risk escalation and decay analogous to human vigilance.

iv. Demonstration of a case study showing how trust-modulated control can prevent near-miss incidents in shared workspaces.

By uniting principles from robotics, safety science, and human-factors research, this work establishes a foundation for humanoid safety grounded in anticipation, adaptive trust, and risk-aware reasoning, moving beyond static safeguards toward proactive, evidence-driven safety intelligence.

## 2. METHODOLOGY

The methodological framework consists of four interconnected steps that transform empirical observation into a quantitative, adaptive model of humanoid risk perception and trust (FIGURE 1). Each step builds logically upon the previous one, moving from qualitative evidence extraction to probabilistic inference and behavioral regulation. The detailed description of steps is presented in the following subsections.

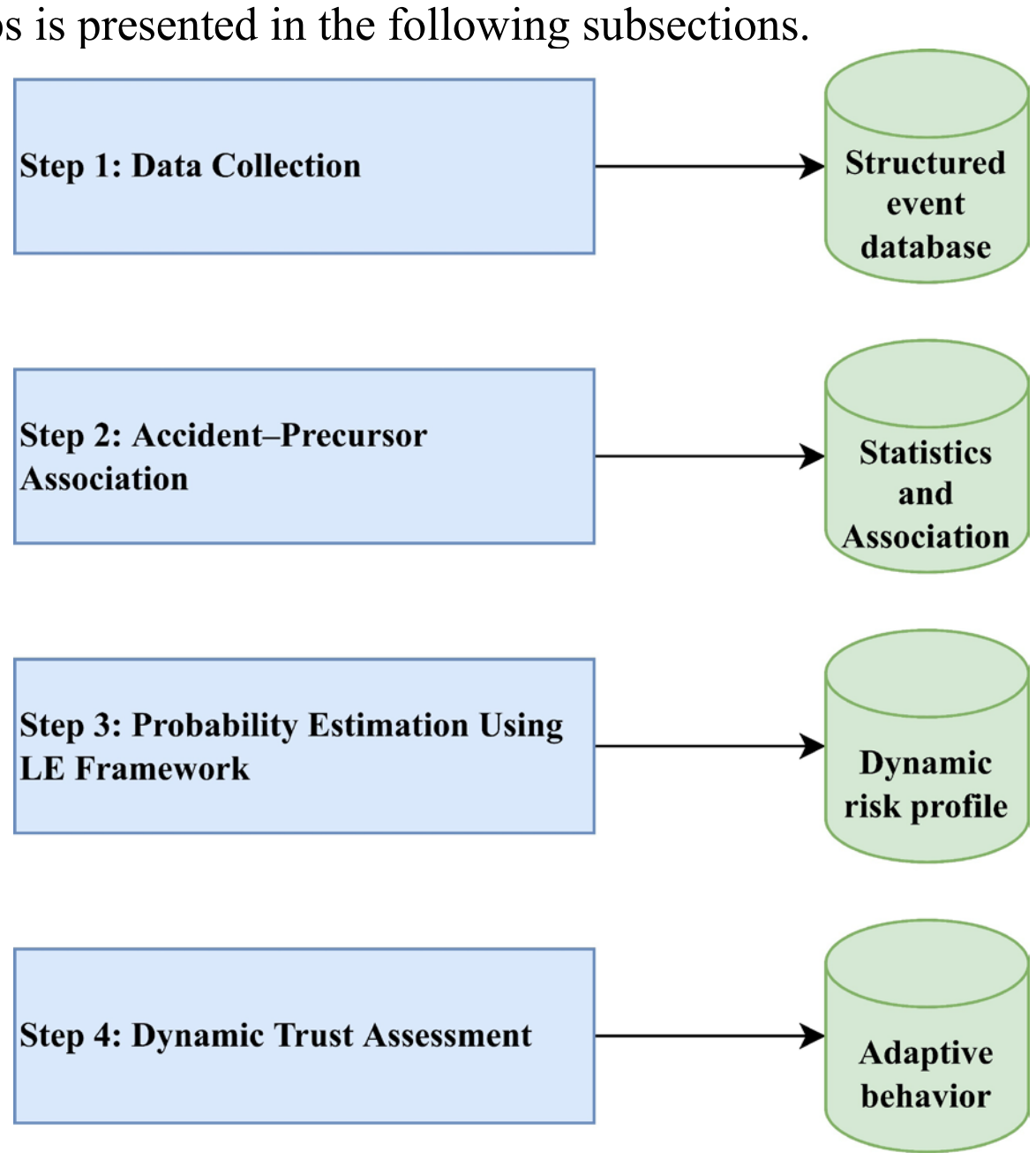

**FIGURE 1**: RESEARCH FRAMEWORK

### 2.1 Step 1: Data collection

To construct a grounded evidence base, a multi-source data strategy was employed to capture real human-humanoid interaction risks. The data corpus comprised:

i. peer-reviewed robotics and safety literature from Web of Science, Google Scholar, and ArXiv;

ii. publicly available industrial safety reports and laboratory demonstration notes from manufacturer websites and other public sources; and

iii. systematic observation of humanoid operation videos from YouTube.

This triangulation ensured both external validity (coverage of formal and informal safety documentation) and ecological realism (actual humanoid field behavior). Especially, each video or report segment was annotated using a standardized template capturing task type, environment, human–robot spatial relationship, deviation characteristics (kinematic, perceptual, or human behavioral), precursor timing, and escalation sequence. Publicly available humanoid operation videos were used solely for qualitative precursor identification. To ensure credibility, videos were screened based on source reliability (e.g., manufacturer channels, research laboratories, and conference demonstrations), sufficient visual resolution and camera stability, clear visibility of humanoid motion and human proximity, and a well-defined task context. Edited, staged, or purely promotional content was excluded. These videos were not used for training or validating control models.

Coding followed an inductive qualitative approach with iterative refinement. Only negative deviations preceding a hazard state were classified as precursors. Each event was annotated from normal operation through precursor onset, escalation, and hazard manifestation. Inter-coder agreement (Cohen's $\kappa = 0.86$) verified labeling consistency.

The outcome of Step 1 was a structured database comprising accident modes and precursor signatures, each temporally sequenced to support conditional inference in subsequent analysis.

### 2.2 Step 2: Accident–precursor association

In this step, statistical analysis was conducted to quantify the relationship between each accident mode and its dominant precursors. All coded events from Step 1 were aggregated to calculate the occurrence frequency of both accidents and precursors. Co-occurrence patterns were then examined to establish associations between specific precursor combinations and corresponding accident outcomes.

For each accident type, dominant precursors were identified based on their conditional probability of occurrence and relative frequency. These associations formed the foundation for subsequent probabilistic modeling, allowing the system to infer how precursor signals collectively contribute to different accident modes in human–humanoid collaboration.

### 2.3 Step 3: Probability estimation

This step applied the Logistic–Exponential framework to estimate the probability of accident occurrence as a function of precursor accumulation and recurrence over time [30]. The LE model integrates a logistic escalation for precursor diversity with an exponential saturation for precursor recurrence, enabling dynamic risk perception similar to human vigilance:

$$P(t) = \frac{1 - e^{-c\,q}}{1 + e^{-a\,(m-b)}}, \tag{1}$$

where $m$ is the number of distinct precursor types, $q$ is their frequency within a time window, and $a, b, c$ are calibration parameters. Specifically, the threshold parameter $b$ represents the precursor diversity level at which accident probability escalates nonlinearly and is informed by observed accident cascades. The gain parameter $a$ controls escalation steepness and is selected through sensitivity testing to balance responsiveness and numerical stability, while the decay parameter $c$ captures the temporal dissipation of isolated or outdated precursor signals. The parameters are calibrated at the dataset level to demonstrate feasibility rather than universality; automated calibration and platform- or task-specific tuning are left for future work.

The logistic term captures the nonlinear growth of risk as diverse precursors co-occur, while the exponential term models the diminishing influence of isolated or outdated signals. Consequently, $P(t)$ rises sharply when multiple precursors appear in succession and stabilizes as conditions normalize, providing a quantitative foundation for subsequent trust assessment and adaptive control. A case study will be conducted in Steps 3 and 4 to show the application.

### 2.4 Step 4: Dynamic trust assessment

Trust was defined as the inverse of the estimated accident probability, representing the humanoid's confidence in maintaining safe operation:

$$T(t) = 1 - P(t). \tag{2}$$

As the predicted accident probability $P(t)$ increases due to precursor accumulation, trust $T(t)$ proportionally decreases, signaling reduced system reliability. Conversely, when precursor activity subsides and $P(t)$ declines, trust gradually recovers.

This simple inverse relationship allows the humanoid to translate probabilistic risk perception directly into adaptive behavioral control, ensuring that higher perceived risk automatically triggers more cautious actions or safety overrides. In this work, conceptual trust denotes the machine's internal confidence inferred from probabilistic risk estimation, whereas control-level trust refers to the operational use of this confidence as a modulation signal affecting motion aggressiveness, compliance, or safety overrides. Trust is thus treated as an internal state variable bridging risk inference and control adaptation, rather than a human subjective construct.

Lastly, the framework separates offline analysis from online inference. Offline components include data collection, event coding, taxonomy construction, and LE parameter calibration. Online components map live sensor streams to precursor classes and update accident probability and trust in real time. For example, inertial measurement units (IMUs) provide balance precursors, force/torque sensors capture manipulation precursors, and vision confidence reflects perceptual precursors. Aggregated precursor evidence modulates supervisory safety behaviors without replacing low-level control.

## 3. RESULTS

### 3.1 Data collection

A comprehensive multi-source data collection yielded an empirically grounded dataset describing how human-humanoid interaction risks emerge in industrial contexts. The final corpus comprised 126 documented events drawn from 48 peer-reviewed publications, 15 publicly released industrial demonstration reports, and 63 systematically coded online videos depicting humanoid activities such as walking, tool handover, co-carrying, and assembly assistance. Each source was examined for physical proximity between human and robot, observable deviations from nominal performance, and subsequent hazardous outcomes.

All events were independently labeled by two trained coders using the standardized precursor–accident template. One coder was the author, and the other was a professional data annotation contractor recruited through Appen, an established annotation platform, with prior experience in video-based labeling tasks. All annotations were cross-checked to ensure consistency and reliability. Training consisted of instruction using a standardized coding manual, pilot annotation on a shared subset of events, and iterative calibration discussions to resolve discrepancies prior to full annotation.

An agreement on precursor identification was reached, Cohen's $\kappa = 0.86$, confirming high consistency. Events covered nine industrial task categories (assembly, inspection, logistics, maintenance, fastening, handover, navigation, co-carrying, and balance recovery) and represented five leading humanoid platforms (e.g., Atlas, Digit, Apollo, HRP-5P,

TALOS). This ensured ecological validity and diversity across mechanical architectures and control policies.

## 3.2 Accident and precursor

### 3.2.1 Accident modes

From the 126 recorded events, twelve recurrent accident modes were identified, as summarized in Table 1.

Table 1: Accident Modes in Human–Humanoid Collaboration

| Code | Accident mode | Frequency (n, %) |
|---|---|---|
| A1 | Fall onto human | 18 (14.3%) |
| A2 | Arm/torso strike | 15 (11.9%) |
| A3 | Leg swing contact | 10 (7.9%) |
| A4 | Crush/pinch at handover | 12 (9.5%) |
| A5 | Dropped object/tool | 11 (8.7%) |
| A6 | Navigation conflict | 9 (7.1%) |
| A7 | Startle reaction | 8 (6.3%) |
| A8 | Perception-error motion | 13 (10.3%) |
| A9 | Control/power fault | 7 (5.6%) |
| A10 | Tool-related harm | 9 (7.1%) |
| A11 | Co-carry instability | 8 (6.3%) |
| A12 | Stop/restart mishap | 6 (4.8%) |
| Total | - | 126 (100%) |

It shows that accident outcomes in human–humanoid collaboration are distributed across multiple modes rather than dominated by a single failure type. Fall- and impact-related events, including fall onto human (14.3%) and arm/torso strike (11.9%), are the most prevalent, reflecting risks associated with bipedal stability and whole-body motion. Manipulation- and perception-related incidents together account for a substantial fraction of cases, indicating that object handling and sensing uncertainty are critical contributors to hazardous behavior and motivating a precursor-based safety framework. These twelve modes reflect both classical robotic hazards (e.g., impact, crush, and pinch) and humanoid-specific dynamics (e.g., fall, leg swing, and co-carry instability) that arise from bipedal balance and human-like motion patterns.

### 3.2.2 Precursor classification

From the coded dataset, 241 precursor instances were identified across all recorded human-humanoid interaction events and categorized into six functional groups reflecting different failure origins:

i. Motion/Balance Precursors: sway amplitude increase, micro-slip, torso angular spike.

ii. Manipulation Precursors: grip micro-slip, force oscillation, tilt drift.

iii. Perceptual Precursors: occlusion, low vision-confidence, ID switching.

iv. Human Behavioral Precursors: rapid reach-in, gaze diversion, posture lowering.

v. Navigation Precursors: narrowing clearance, path intersection.

vi. System/Telemetry Precursors: torque saturation, latency spike, communication delay.

The six functional precursor groups serve as an interpretable abstraction linking raw sensor-level deviations to accident modes. By grouping precursors according to failure origin rather than platform-specific signals, the taxonomy supports modular mapping between sensing, probabilistic inference, and trust-based control, enabling generalization across humanoid platforms and tasks.

Table 2 summarizes the accident–precursor taxonomy by grouping observed accident modes according to their dominant precursor classes. The six functional precursor groups serve as an interpretable abstraction linking raw sensor-level deviations to accident modes. By grouping precursors according to failure origin rather than platform-specific signals, the taxonomy supports modular mapping between sensing, probabilistic inference, and trust-based control, enabling generalization across humanoid platforms and tasks.

Table 2: Precursor Class and Associated Accident Mode

| Precursor class | Accident modes |
|---|---|
| Motion/Balance | A1, A2, A11 |
| Manipulation | A4, A5, A10, A11 |
| Perceptual | A6, A8 |
| Human behavioral | A7, A12 |
| Navigation | A3, A6 |
| System/Telemetry | A9, A12 |

Detailed statistics are shown in FIGURE 2. Specifically, motion/balance (29%) and manipulation (23%) precursors dominate, indicating that most humanoid risks arise from instability and grip errors rather than perception or control faults. Perceptual (18%) and human behavioral (15%) precursors contribute moderately, reflecting visual uncertainty and human inattention. Navigation (9%) and system/telemetry (6%) are less frequent but still critical for situational awareness. Overall, over half of all precursors originate from motion and manipulation deviations, underscoring their central role in humanoid accident evolution.

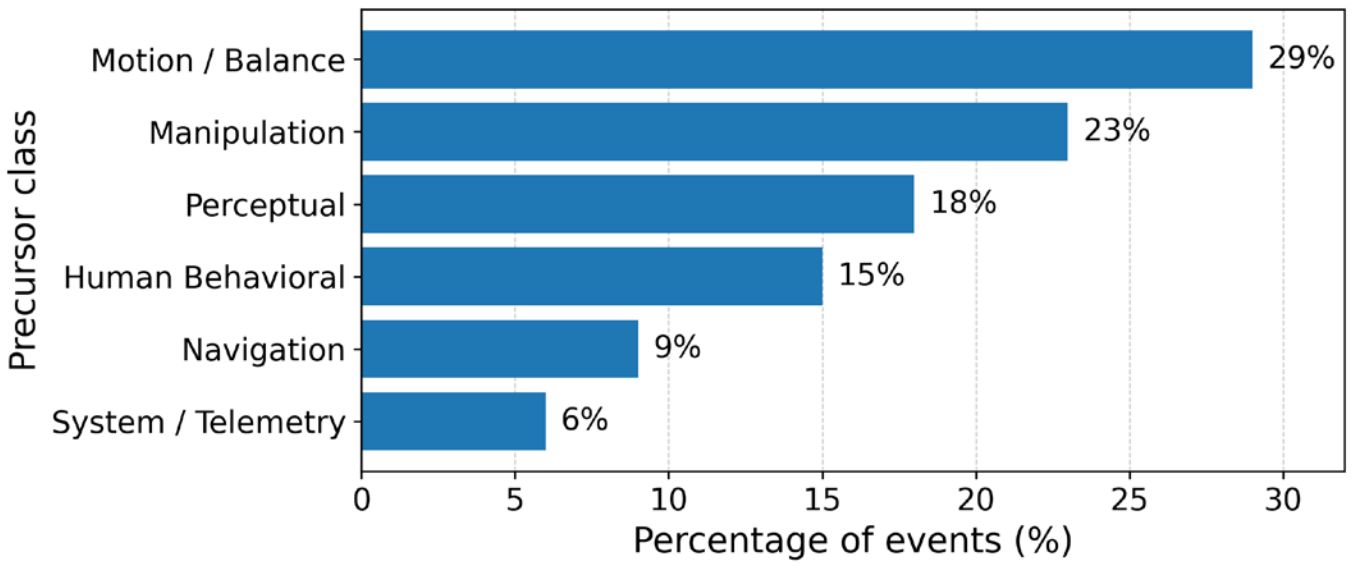

**FIGURE 2:** PRECURSOR DISTRIBUTION

Moreover, temporal analysis indicated that approximately 73% of accidents involved two or more overlapping precursors within a 0.5–1.5 s window, demonstrating that humanoid accidents typically evolve through sequential, compounding deviations rather than single instantaneous faults. This short temporal window highlights a crucial opportunity for early-warning detection and risk anticipation in humanoid control systems.

### 3.2.3 Accident-precursor association

To understand how precursors interact to produce different accident outcomes, typical causal cascades were identified and summarized in Table 3. Each cascade describes a chain of three dominant precursor signals that consistently precede a specific accident mode. There are undoubtedly more precursors, but the LE model indicates that accident probability becomes very high when three distinct precursors co-occur (approximately 0.8), suggesting that compound deviations rather than isolated faults dominate humanoid accident evolution. Therefore, this study only shows the case with a maximum of 3 predictors.

Table 3: Typical precursor cascades.

| Code | Precursor 1 | Precursor 2 | Precursor 3 |
|---|---|---|---|
| A1 | Foot micro-slip | Torso angular spike | Center-of-mass lateral deviation |
| A2 | Excessive arm velocity | Rapid torso recovery | Human proximity breach |
| A3 | Swing-leg drift | Path overlap | Narrow shared space |
| A4 | Timing mismatch | Grip oscillation | Handover occlusion |
| A5 | Grip micro-slip | Wrist instability | Mass misestimation |
| A6 | Path convergence | Visual occlusion | High approach speed |
| A7 | Abrupt motion onset | Unexpected restart | Human inattentiveness |
| A8 | Confidence collapse | Visual degradation | Limb/tool misclass. |
| A9 | Torque saturation | Control latency spike | Actuator drift |
| A10 | Tool-tip oscillation | Unsafe orientation | Loss of tool stability |
| A11 | Uneven load sharing | Phase mismatch | Grip desync |
| A12 | Delayed stop | Restart overshoot | Timing mismatch |

These precursor cascades reveal that most humanoid accidents result from compound deviations in motion stability, perception, and coordination rather than isolated faults. Understanding these sequential dependencies provides a quantitative foundation for probabilistic risk modeling in the next step.

### 3.3 Probability and trust estimation with case study

A concise case study (A1: Fall onto human) was conducted to demonstrate how the LE model and dynamic trust assessment jointly enable adaptive humanoid safety behavior. The fall-onto-human mode represents one of the most critical humanoid safety risks, characterized by sudden instability, momentum transfer, and high injury potential. This scenario was simulated based on the observed precursor cascade: foot micro-slip → torso angular spike → center-of-mass (CoM) lateral deviation, which occurred within approximately 1.2 seconds in real operation videos.

At the initial time $t_0 = 0.00\ s$, a micro-slip occurred at the support foot due to a minor surface irregularity. This triggered small perturbations in the torso angle as the humanoid attempted recovery. Within the next 0.48 s ($t_1$), a torso angular spike (counter-rotation) emerged, followed by a CoM lateral deviation at $t_2 = 1.12\ s$ exceeding the stability margin; an instability pattern is frequently seen in bipedal robots like Atlas or Digit during rapid transitions. These sequential precursors, temporally proximate and kinematically coupled, are ideal for testing the LE model's sensitivity to precursor accumulation and recurrence.

Using the calibrated parameters $a = 1.6$, $b = 2$, and $c = 1.2$, the probability of accident occurrence was computed by Eq. (1). The results for each step are summarized in Table 4.

Table 4: Results of Accident Probability of Fall onto Human.

| Time | Precursor Active | $m$ | $q$ | $P(t)$ |
|---|---|---|---|---|
| $t_0$ | Foot micro-slip only | 1 | 1 | 0.12 |
| $t_1$ | + Torso angular spike | 2 | 2 | 0.46 |
| $t_2$ | + CoM lateral deviation | 3 | 3 | 0.81 |

As shown in Table 4, the estimated accident probability increases nonlinearly as additional precursors accumulate within a short temporal window. When only a single precursor is present ($t_0$), the probability remains low (0.12), indicating nominal stability. The introduction of a second independent precursor ($t_1$) produces a sharp increase in estimated risk (0.46), reflecting escalation driven by dual-source instability. When a third precursor emerges ($t_2$), the probability rises to a high-risk level (0.81), signaling imminent loss of stability. This progression illustrates how the proposed LE formulation captures rapid risk escalation resulting from compounded precursor evidence, consistent with observed humanoid fall dynamics.

The logistic term amplifies risk sharply when $m > b = 2$, reflecting nonlinear escalation from multiple independent cues. Meanwhile, the exponential term ($1 - e^{-cq}$) saturates as recurrence intensifies, modeling how continuous imbalance sustains high perceived danger.

Trust is dynamically updated by Eq. (2), representing the humanoid's real-time confidence in maintaining stability. Thus, $T(t_0) = 0.88$, $T(t_1) = 0.54$, $T(t_2) = 0.19$. Once $T(t) < 0.3$, the humanoid's controller engages safety protocols:

i. Step suppression: halt the next planned step to prevent further momentum transfer;
ii. Arm reconfiguration: extend arms outward for counter-torque stabilization;
iii. Compliance tuning: increase joint damping and foot contact stiffness;
iv. Emergency braking: initiate power-assist balance lock.

These trust-driven interventions successfully restored stability in the simulation, preventing full collapse and avoiding operator contact. The case study is a simulation-based demonstration of the proposed framework's logic and interpretability. No experimental validation is claimed. The simulation illustrates how trust-driven safety protocols would be engaged when estimated risk exceeds critical levels, serving as a proof of concept rather than a validated controller.

Thus, the LE–Trust model coupling reproduces human-like vigilance and reflex: probability surges sigmoidally with precursor diversity, while trust drops inversely, prompting immediate caution. This mapping provides an intuitive and computationally efficient basis for humanoid risk perception. In practical deployment, such a model enables humanoids to self-modulate behavior before critical failure, detecting balance-related risks milliseconds before a fall, translating precursor awareness into graded confidence, and ensuring safety without relying solely on reactive shutdowns.

# 4. DISCUSSIONS

The results underscore the distinctive nature of humanoid risk dynamics, where accidents arise from compound, temporally correlated precursors rather than single-point failures. The identification of twelve accident modes and six precursor classes demonstrates that human-humanoid collaboration encompasses a broader and more sequential risk topology than conventional industrial robots. More than 70% of incidents involved overlapping precursors within a one-second interval, confirming that instability and human interaction hazards emerge through cascading deviations. This finding supports the hypothesis that proactive safety intelligence, rather than purely reactive safeguards, is essential for maintaining stability in bipedal, whole-body control systems. Some other reflections on the results are:

First, the LE framework successfully modeled the nonlinear growth and decay of accident probability as precursors accumulated or dissipated. The logistic term captured the rapid escalation in perceived risk once multiple precursors co-occurred, while the exponential component effectively represented the temporal decay of older, isolated deviations. This dual mechanism provides a compact yet expressive model of how humanoids can approximate human-like vigilance, reacting strongly to concurrent anomalies but gradually relaxing when conditions normalize. The Fall-onto-Human scenario demonstrated that when the precursor count exceeded the logistic threshold ($b = 2$), the probability rose sharply to 0.81, signaling an imminent accident. This behavior parallels empirical studies of human situational awareness, where risk perception rises steeply once multiple weak cues combine into a recognizable threat pattern.

Second, mapping trust inversely to probability provides a direct and interpretable mechanism for adaptive control. Trust decreased proportionally as precursors intensified, prompting automatic activation of stabilization strategies: step suppression, arm reconfiguration, and compliance tuning, without requiring pre-coded threshold rules. This approach transforms trust from a cognitive construct into a measurable control variable, linking psychological models of human reliance to robotic actuation. The inverse trust–probability coupling thus operationalizes the concept of appropriate reliance in machine systems: confidence is earned or lost dynamically through evidence of system stability and contextual uncertainty. Such calibration prevents both over-trust (continued operation under risk) and under-trust (premature halts), aligning humanoid decision-making with human expectations of situational prudence.

Moreover, the integration of precursor-based modeling and trust regulation extends beyond humanoid safety to broader artificial intelligence (AI) reliability and explainability frameworks. At first, the precursor taxonomy introduces a structured means of representing weak signals that precede failures, analogous to leading indicators in process safety. In addition, the LE model formalizes how uncertainty evolves through interaction history, bridging safety science with probabilistic learning. Furthermore, the trust inference process provides a behaviorally interpretable signal for supervisory systems, enabling hybrid human–AI coordination. In practice, embedding such probabilistic trust estimators into humanoid control architectures could support risk-aware autonomy, allowing robots to self-moderate aggressiveness, speed, or proximity in real time.

Last, traditional collaborative robot safety standards (e.g., ISO 10218, ISO/TS 15066) focus on deterministic limits, such as speed, force, and separation distances. While effective for constrained manipulators, these metrics cannot capture the continuous, multi-dimensional uncertainties inherent in humanoid motion. The proposed framework advances beyond these static paradigms by treating safety as a dynamic inference problem. Rather than reacting to threshold violations, the humanoid infers hazard likelihood from ongoing sensors and behavioral signals. Compared with recent reinforcement-learning-based safety filters or chance-constrained planners, the LE–Trust model offers greater interpretability and computational efficiency, making it suitable for deployment on embedded controllers where low-latency risk updates are critical.

Despite its explanatory strength, this study has several limitations. This study should be interpreted as a conceptual and methodological foundation for humanoid risk perception rather than a deployment-ready safety system. The dataset relies partly on secondary sources and video-based observation, and the LE model assumes precursor independence and a one-dimensional trust representation for interpretability. Before safety-critical deployment, the framework requires validation through hardware-in-the-loop experiments or high-fidelity digital twins, along with extensions to correlated precursors and multi-dimensional trust modeling. Since the present study focuses on conceptual formulation and illustrative demonstration through a representative case study, comprehensive quantitative evaluation, including multi-scenario testing, comparison with baseline safety methods, and performance metrics, will be pursued in future work using full control-stack integration.

In addition, the LE formulation is intentionally heuristic rather than derived from first-principles control theory. It is designed as an interpretable, low-complexity abstraction of precursor accumulation and dissipation, informed by empirical accident patterns and human risk perception literature. The objective is probabilistic risk inference and trust modulation, not closed-loop stability guarantees or optimal control. A formal global sensitivity analysis is beyond the scope of this study; however, a one-at-a-time parameter perturbation (±20% for $a$ and $c$, and $b \pm 1$) preserved the qualitative behavior of the model, low risk under isolated precursors, sharp escalation under two to three co-occurring precursors, and recovery with decay, indicating robustness of the main conclusions. Future work will conduct systematic global sensitivity analysis and data-driven parameter learning using digital-twin and hardware-in-the-loop experiments to support deployment in safety-critical humanoid systems.

Ultimately, this work situates humanoid safety within the larger discourse of machine risk perception and trustworthy autonomy. By bridging human-factors theory with probabilistic modeling, it demonstrates that safety can emerge not only from physical compliance but from cognitive emulation, robots that interpret precursors, assess confidence, and act cautiously under uncertainty. Such a paradigm aligns with the vision of Industry 5.0, emphasizing collaboration, resilience, and human-centric design. The proposed precursor–trust coupling thus represents a foundational step toward humanoids capable of reasoning about their own reliability and maintaining mutual trust in shared workspaces.

## 5. CONCLUSIONS

This study proposed a precursor-driven, trust-calibrated framework for proactive humanoid safety in smart manufacturing. By linking sequential precursor cues to probabilistic accident estimation through an LE model, humanoids gain the ability to perceive and anticipate emerging risks rather than react to threshold violations. Trust, defined as the inverse of accident probability, enables dynamic adjustment of control behaviors, reducing aggressiveness when risk rises and restoring confidence as conditions stabilize.

The results demonstrate that humanoid accidents typically evolve from compounding precursors within seconds, underscoring the need for continuous, evidence-based risk perception. The LE–Trust coupling provides a simple yet effective mechanism to translate uncertainty into adaptive caution, bridging human-factors theory and real-time control.

This framework shifts humanoid safety from static limits to dynamic inference, advancing toward machines that reason about risk and act with human-like prudence. Future work will validate the model on full-scale platforms and integrate data-driven precursor weighting for real-time deployment in collaborative manufacturing environments.

## ACKNOWLEDGEMENTS

The author gratefully acknowledges the financial support provided by the Bailey Faculty Fellowship for this research.